\documentclass[5p]{elsarticle}

\usepackage{hyperref}
\usepackage{lineno}

\usepackage{graphicx}
\usepackage{subfigure}
\usepackage{amsmath,amssymb}
\usepackage{multirow}
\usepackage{algorithm}
\usepackage[noend]{algorithmic}

\usepackage[utf8]{inputenc}

\usepackage{color}
\usepackage{soul}

\journal{Knowledge-Based Systems}

\begin{document}
	
	\begin{frontmatter}
		
		\title{Generalized multiscale feature extraction for remaining useful life prediction of bearings with generative adversarial networks}
		
		\author[DFKI]{Sungho Suh \fnref{suh}}
		\author[DFKI]{Paul Lukowicz}
		\author[KISTEurope]{Yong Oh Lee\corref{mycorrespondingauthor}}
		\cortext[mycorrespondingauthor]{Corresponding author}
		\ead{yongoh.lee@kist-europe.de}
		
		\fntext[suh]{This work was done while the author was at KIST Europe.}
		\address[DFKI]{German Research Center for Artificial Intelligence (DFKI), 67663 Kaiserslautern, Germany}
		\address[KISTEurope]{Smart Convergence Group, Korea Institute of Science and Technology Europe Forschungsgesellschaft mbH, 66123 Saarbrücken, Germany}

		\begin{abstract}
			Bearing is a key component in industrial machinery and its failure may lead to unwanted downtime and economic loss. Hence, it is necessary to predict the remaining useful life (RUL) of bearings. Conventional data-driven approaches of RUL prediction require expert domain knowledge for manual feature extraction and may suffer from data distribution discrepancy between training and test data. In this study, we propose a novel generalized multiscale feature extraction method with generative adversarial networks. The adversarial training learns the distribution of training data from different bearings and is introduced for health stage division and RUL prediction. To capture the sequence feature from a one-dimensional vibration signal, we adapt a U-Net architecture that reconstructs features to process them with multiscale layers in the generator of the adversarial network. To validate the proposed method, comprehensive experiments on two rotating machinery datasets have been conducted to predict the RUL. The experimental results show that the proposed feature extraction method can effectively predict the RUL and outperforms the conventional RUL prediction approaches based on deep neural networks.
		\end{abstract}
		
		
	\end{frontmatter}
	
	
	\section{Introduction}
	\label{introduction}
	
	
	Prognostics and health management (PHM) technology collects status information from industrial systems, such as manufacturing machines, facilities, and power plants, to detect failures of the system and enables maintenance schedule in advance by predicting the point of failure through analysis and predictive verification \cite{vachtsevanos2006intelligent}. One of the PHM technologies predicts the remaining useful life (RUL) of the rolling element bearings to prevent unexpected failures and improve reliability \cite{li2015improved, qian2017multi}. A large number of studies have been conducted for RUL prediction, which can be categorized into model-based \cite{pecht2009physics} and data-driven methods \cite{guo2016hierarchical}. Model-based methods estimate the RUL through analytical models based on physical laws and mathematical functions. These methods include physics-based methods \cite{pecht2009physics}, empirical-based methods \cite{vachtsevanos2006intelligent}, Kalman filter \cite{ompusunggu2015kalman}, and particle filter \cite{jouin2016particle}. 
	However, they require expert domain knowledge to build a precise model in an increasingly complex industrial systems \cite{chen2012machine}.	
	Recently, data-driven methods have garnered significant interest along with substantial development in machine learning and minimal requirement of expert knowledge. The data-driven methods capture the direct relationship between collected machine data and degradation status with machine learning techniques. 
		
	Recently, deep-learning-based RUL prediction methods have been proposed and have achieved better prediction performance than conventional data-driven methods \cite{ren2017multi, guo2017recurrent, ren2018bearing, hinchi2018rolling, zhu2018estimation, li2019deep, ren2019multi, qin2020gated}. Although these deep-learning-based RUL prediction methods have been successfully developed, less importance has been given to three challenging issues: (1) These methods require domain knowledge to extract features or we have to manually specify the feature types. 
	(2) The aforementioned data-driven methods assume that training and test data are collected by the same sensors under the same operating conditions or are from the same distribution. However, these assumptions can be impractical in industries, since machinery working conditions usually change with respect to tasks and the training and test data can be collected from different entities. 	 
	(3) For accurate RUL prediction, it is essential to properly determine the health stage (HS) of the machinery; because the machine in a healthy state would not have any remarkable difference in the run-to-failure training dataset. However, extant conventional methods predict the RUL without determining the first predicting time (FPT) \cite{da2020remaining, ragab2020adversarial, ragab2020contrasive}, which is the start time of the unhealthy stage. 
	
	In this study, we propose a generalized multiscale feature extraction method for the RUL prediction. Generative adversarial network(GAN) is used to learn the distributions of multiple training data from different bearings and extract domain-invariant generalized prognostic features. For the determination of FPT and RUL prediction, the proposed feature extraction method consists of two steps; the first of which trains multiscale adversarial neural networks to reconstruct the vibration input signals to the generalized prognostic features. Here, three different levels of a one-dimensional U-Net \cite{ronneberger2015u} architecture are trained to minimize the loss functions of the GAN-based proposed feature extraction method; the second step transforms the generalized features into a nested-scatter plot (NSP) \cite{suh2019generative} image to determine the FPT and to predict the RUL. 
    NSP is an imaging method for multi-variable correlation analysis. NSP is heuristic, but the NSP images transformed from raw vibration signals reduce the efforts for feature engineering based on domain knowledge. Also, NSP images are useful for the feature extraction in fault diagnosis of rotating machinery, when combined with convolutional neural networks (CNN) \cite{suh2019generative}. A CNN-based binary regression model determines the FPT, and a CNN-long short-term memory (CNN-LSTM) model predicts the RUL.
	
	The main contributions of this work are summarized as follows.
	
	\begin{itemize}[]
		\item A novel multiscale feature extraction method designed for the HS division and the RUL prediction. We formulate one-dimensional feature extraction as a principal signal separation task and introduce the use of U-Net to reconstruct the prognostic features for the RUL prediction. A novel domain-invariant generalized solution based on the GAN scheme is introduced to learn the invariant representation and predict the RUL.
		\item We transform the multiscale prognostic features into an NSP image without any domain knowledge and manual setting to combine the generalized multiscale features and reduce the computational cost.
		\item A CNN-based binary regression model for determination of the HS without any threshold and a CNN-LSTM model for the RUL prediction are proposed, which can predict the RUL with less errors and higher prognosis accuracy than other existing methods.
		\item To validate the proposed method, we conducted experiments with two rotating machinery datasets: the Fanche-Comte Electronics Mechanics Thermal Science and Optics—Sciences and Technologies Institute (FEMTO) dataset \cite{javed2013feature} and the Xi'an Jiaotong University and the Changxing Sumyoung Technology Company (XJTU-SY) dataset \cite{wang2018hybrid}. By the experiments on multiple datasets, we can verify the effectiveness of the proposed method on different patterns of bearing wear.
	\end{itemize}
	
	The rest of the paper is organized as follows. Section \ref{sec:relatedworks} introduces the related works. Section \ref{sec:proposedmethod} provides the details of the proposed method. Section \ref{sec:experimentalresults} presents quantitative and qualitative experimental results on the two datasets. Finally, Section \ref{sec:conclusion} concludes the paper.

	\section{Related works}
	\label{sec:relatedworks}
	\subsection{Deep Learning and Adversarial Domain Adaptation for RUL}
		
	Ren et al. \cite{ren2017multi} proposed an integrated deep neural network (DNN) that extracts both the time and frequency-domain features of vibration signals and predicts the bearing RUL. Guo et al. \cite{guo2017recurrent} employed a recurrent neural network (RNN) based on health indicator to predict the RUL of bearings, where six related-similarity features with eight time-frequency features were fed into the recurrent neural networks(RNN). Ren et al. \cite{ren2018bearing} proposed a deep learning-based RUL prediction method using a deep auto-encoder (DAE) and DNN. Hinch et al. \cite{hinchi2018rolling} proposed an end-to-end RUL prediction method based on convolutional and LSTM neural networks by taking raw data as input. Zhu et al. \cite{zhu2018estimation} proposed a multiscale CNN-based RUL prediction method with the time-frequency representation using the wavelet transformation. Li et al. \cite{li2019deep} proposed a deep learning-based RUL estimation method with multiscale feature extraction and FPT determination from the kurtosis of vibration signals. Ren et al. \cite{ren2019multi} proposed a deep learning method using the multiscale dense gated RNN for the RUL prediction. Qin et al. \cite{qin2020gated} proposed a gated dual attention unit for the RUL prediction. 
	
	These works have achieved satisfactory performance in RUL prediction, nonetheless the three challenges still remain a concern. Firstly, these works require domain knowledge to extract features or we have to manually specify the feature types. The various time-domain statistical features, such as signal root mean square (RMS), crest factor, kurtosis, and entropy, and the frequency-domain features, such as fast Fourier transform (FFT), Hilbert-Huang transform (HHT), and wavelet transformation, were extracted and used to predict the RUL \cite{ren2017multi, guo2017recurrent, ren2018bearing, ren2019multi}. In other words, only one time-domain feature or time-frequency feature by wavelet transformation \cite{zhu2018estimation}, short-time Fourier transform (STFT) \cite{li2019deep}, or RMS \cite{qin2020gated} was used to predict the RUL. The RUL prediction can remove the feature extraction by simply taking raw data as input to the deep neural networks (DNNs) \cite{hinchi2018rolling}, but it results in high computation cost. Even worse, such a method cannot correct the training error, if the extracted feature is used to predict RUL with the long short-term memory (LSTM) neural networks. Secondly, though the machinery working conditions, such as motor power and rotating speed, and degradation generation conditions, such as radial force, are identical in an experimental platform to validate bearings fault detection, the bearing faults usually vary. Generally, the degradation trends are not the same for different bearings under the same conditions, resulting in significant data distribution discrepancies between entities. To overcome these problems, adversarial domain adaptation approaches and CNN-based health stage prediction methods have been proposed for RUL prediction \cite{da2020remaining, ragab2020adversarial, ragab2020contrasive, li2020data}. The adversarial domain adaptation approaches have been proposed to generalize the feature extraction of bearing wear. Da Costa et al. \cite{da2020remaining} proposed a deep domain adaptation method for the RUL prediction by integrating the LSTM model with domain classification loss to reduce the deviation between source and target domains. Ragab et al. \cite{ragab2020adversarial, ragab2020contrasive} proposed a contrastive adversarial domain adaptation approach that enables automatic feature extraction and the RUL prediction from a source domain to a target domain. However, these approaches predict the RUL without determining the HS and FPT and only focus on the single-source single-target adaptation setting which is impractical because the network of domain adaptation should be trained for data from different domains with different distributions. However, those approaches only focus on the single-source single-target adaptation setting. In real-world applications, the training data may come from multiple domains with different distributions so that it is expensive in both time and cost to train the network of domain adaptation for every condition and bearing.	

	 Thirdly, the conventional methods predict RUL without determining the FPT. The FPT can be determined using conventional features, such as kurtosis and RMS \cite{li2019deep, qin2020gated}. Although unsupervised methods \cite{principi2019unsupervised}, such as auto-encoder, promise the extraction of efficient health indicators(HI); however, they possess a challenge assigning an appropriate threshold for determining FPT. Recently, the CNN-based HS division methods have been proposed, but they require an appropriate threshold to determine the FPT \cite{li2020data} or need to determine bandwidths, where there are significant differences between normal and faulty conditions \cite{suh2019generative, suh2020supervised}. Li et al. \cite{li2020data} proposed a GAN-based RUL prediction approach that consists of two stages. Initially, the GAN is used to learn the distributions of the data in the healthy state of the machine and determine the FPT in prognostics. The learned features are further adopted for the RUL prediction and adversarial training is also utilized for data alignments of different entities. It trains the DNN with the degradation data from multiple training data of different bearings and the healthy state data from different training and test bearings and can generalize prognostic features. However, it still requires an appropriate threshold to determine the FPT and needs the target domain data to train the DNN. 
	 Our previous works \cite{suh2019generative, suh2020supervised} proposed a supervised HS division method using CNN and an NSP, which transforms the time-series data into an image for feature extraction, rids the necessity of designating a threshold of health index. However, it still needs to determine bandwidths of bandpass filters manually in the data preprocessing procedure. Also, manual bandpass filter setting based on the sampled dataset can be limited, because it is hard to consider all machinery working conditions.
	 
	 \subsection{Nested Scatter Plot (NSP)}
	 NSP is a data wrangling method that transforms correlated time series data into an image for multi-variate correlation analysis \cite{suh2019generative, suh2020supervised}. It is a matrix representation similar to the heat map of the quantized value of time series data. Each vibration signal is quantized and mapped into nested clusters. The cumulative number of signal values in the nested cluster is normalized to represent the density of the nested cluster. Although NSP removes the non-stationarity of the time sequence, it is an efficient imaging method for multi-variable correlation analysis \cite{suh2019generative}.
	 
	 By using NSP, we transform multi-channel vibration signals into a single fixed-size image. Continuous multi-channel vibration signals are split into  given intervals. A three-step approach is used: feature extraction using bandpass filters, decomposition of the nested clusters in each bandwidth, and aggregation of the decomposed sections into a single RGB image. At least two different data channels as data sources are required and the first step is the incorporation extraction of signals in three different bandwidths. HHT and FFT are used to determine the bandwidth of bandpass filters \cite{suh2019generative}. In the second step, the extracted two-channel signals are compressed into nested clusters. Each channel signal is mapped on the x-axis and y-axis. Three different extracted signals are colored in red, green, and blue, respectively. In the final stage, three scatter plots are aggregated together to form a single RGB image. 
	 
	 In the previous works \cite{suh2019generative, suh2020supervised}, the combination of NSP, which allowed us to integrate a frequency analysis that helped to extract faulty wear features, and a CNN-based binary regression model, rid the necessity of designating a threshold, one of the most crucial problems in early fault detection. The CNN based model effectively distinguished healthy and unhealthy states unambiguously by changing the anomaly detection problem into a binary regression problem, utilizing a simple trigger mechanism. To minimize the labeling process in supervised learning, a small portion of the dataset was utilized for training the model.

	\section{Proposed method}
	\label{sec:proposedmethod}
	
	\begin{figure*}
		\centering
		\includegraphics[width=2\columnwidth]{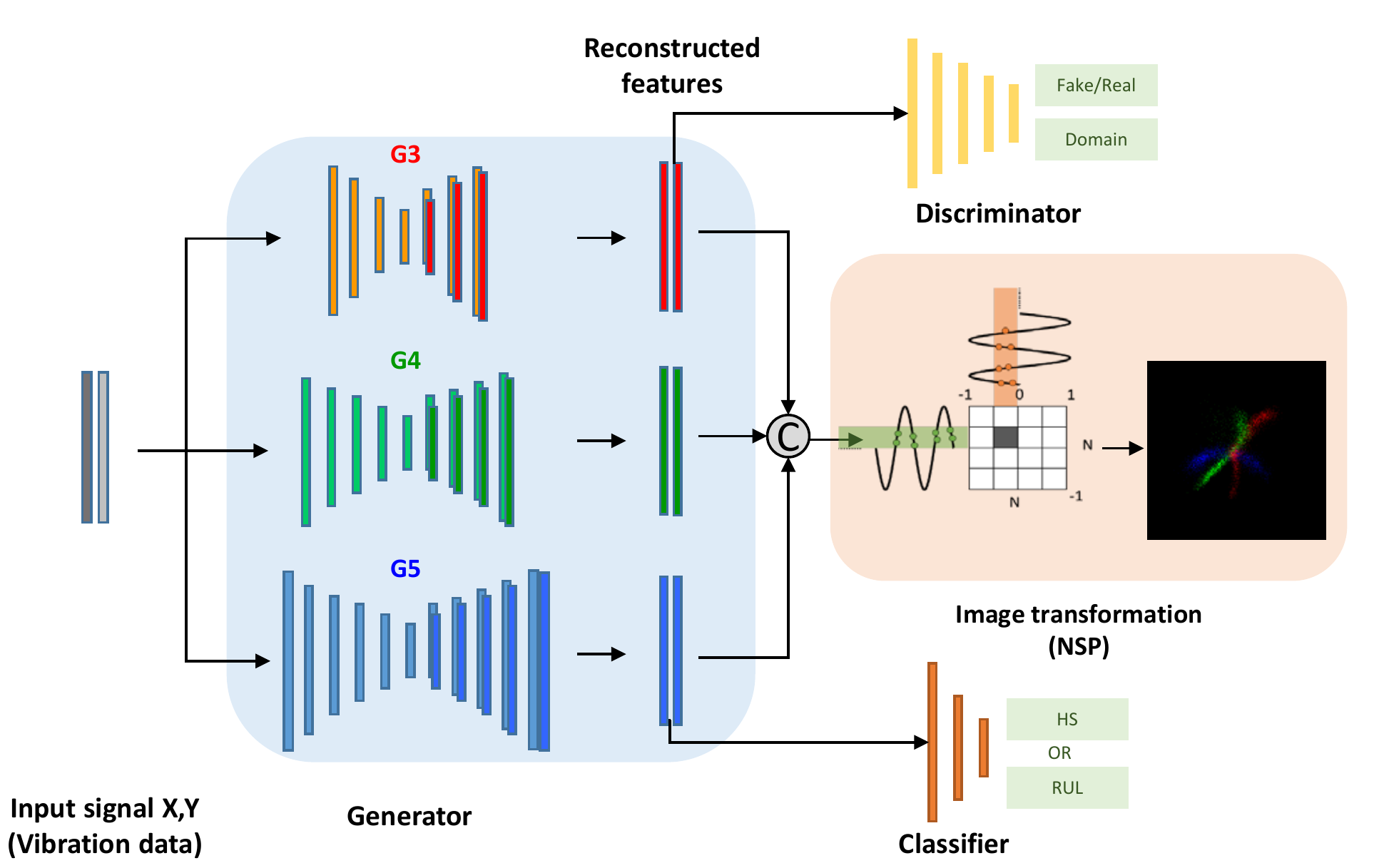}
		\caption{The structure of the proposed framework for the generalized multiscale feature extraction.}
		\label{fig:fig_overview}
	\end{figure*}
	
	\subsection{Generalized Multiscale Feature Extraction}
	
	In this section, we introduce a generalized multiscale feature extraction method based on GAN scheme and U-Net architecture for the HS division and the RUL prediction. In the original formulation of GAN \cite{goodfellow2014generative}, a GAN model is trained through the min-max game between a generator network $G$ and a discriminator $D$. The GAN aims to approximate the probability distribution function that certain data is assumed to be drawn from. The objective function of the min-max game between the generator and the discriminator is expressed as follows:
	
	\begin{equation}
	\label{eq:gan}
	\begin{split}
	\min_{G} \max_{D} \mathop{\mathbb{E}_{x\sim P_r(x)}}[\log D(x)] + \mathop{\mathbb{E}_{z\sim P_z(z)}}[\log (1-D(G(z)))],
	\end{split}
	\end{equation}
	where $x$ is real data sampled from the real data distribution $P_r(x)$, $z$ is the noise vector sampled from a uniform distribution $P_z(z)$, and the generator $G$ generates a synthetic data $G(z)$. The equation shows that treating the discriminator as a classifier minimizes the Jensen-Shannon (JS) divergence between the real data distribution and the one assumed by the generator. 
	
	Although GAN can generate synthetic data close to the real samples, the GAN training procedure has instability with respect to loss function convergence. To solve this instability problem, the Wasserstein GAN with gradient penalty (WGAN-GP) \cite{gulrajani2017improved}, which uses the Wasserstein-K distance as the loss function, was proposed to guide the training process. The loss function of the WGAN-GP is defined as follows.
	
	\begin{equation}
	\label{eq:wgangp}
	\begin{split}
	\mathop{\mathbb{L}} = &- \mathop{\mathbb{E}_{x\sim P_r(x)}}[D(x)] + \mathop{\mathbb{E}_{z\sim P_z(z)}}[D(G_\theta(z))] \\
	&+ \alpha \mathop{\mathbb{E}_{\hat{x}\sim P_{\hat{x}}}}[(\Arrowvert \nabla_{\hat{x}} D(\hat{x}) \Arrowvert_2 -1 )^2],
	\end{split}
	\end{equation}
	where $\alpha$ is the penalty coefficient and $P_{\hat{x}}$ is the uniform sampling along straight lines between pairs of points from the real data distribution $P_r$ and the generated distribution. The motivation for this is that the constraint is enforced uniformly along the line as the optimal discriminator consists of straight lines connecting the two distributions. The WGAN-GP provides a training procedure that is faster and more stable than the original GAN.

	Figure \ref{fig:fig_overview} presents the overall framework with the detailed steps of the learning procedure of the proposed feature extraction model. The GAN structure is a kind of adaptation of our previous work, classification enhancement GAN (CEGAN) \cite{suh2021cegan, suh2022discriminative}. CEGAN is composed of three independent networks: a discriminator, a generator, and a classifier. The classifier in CEGAN is trained with both real and generated data for preventing generated minority data from overfitting to majority data in the imbalanced data, unlike conventional GAN methods. In other words, the generated data for minority classes can prevent the classifier from overfitting to the majority data and improve the performance of the classifier, while the classifier only trained with real data can be biased to the majority class data.
	Because the conventional GAN methods employ the auxiliary classifier and the classifier shares network structure and weight parameters with the discriminator, the performance of the auxiliary classifier cannot lead to generating high-quality images. To generate data to improve the performance of the classifier, CEGAN employs the structure of the classifier to the independent network in the GAN structure. As the structure of the classifier is employed in the GAN structure, the GAN can generate data to improve the performance of the classifier. 	
	
	By adapting the structure of CEGAN, we define three types of independent networks in the proposed GAN scheme: 1) the multiscale feature extractor as a generator network, 2) a discriminator to separate the features generated by the generator from the real input data and to distinguish the data from different domains, 3) the CNN-based HS division and the LSTM-based RUL predictor as the classifier network.

	\subsubsection{Generator}
		
\begin{figure}[!t]
		\centering
		\includegraphics[width=\columnwidth]{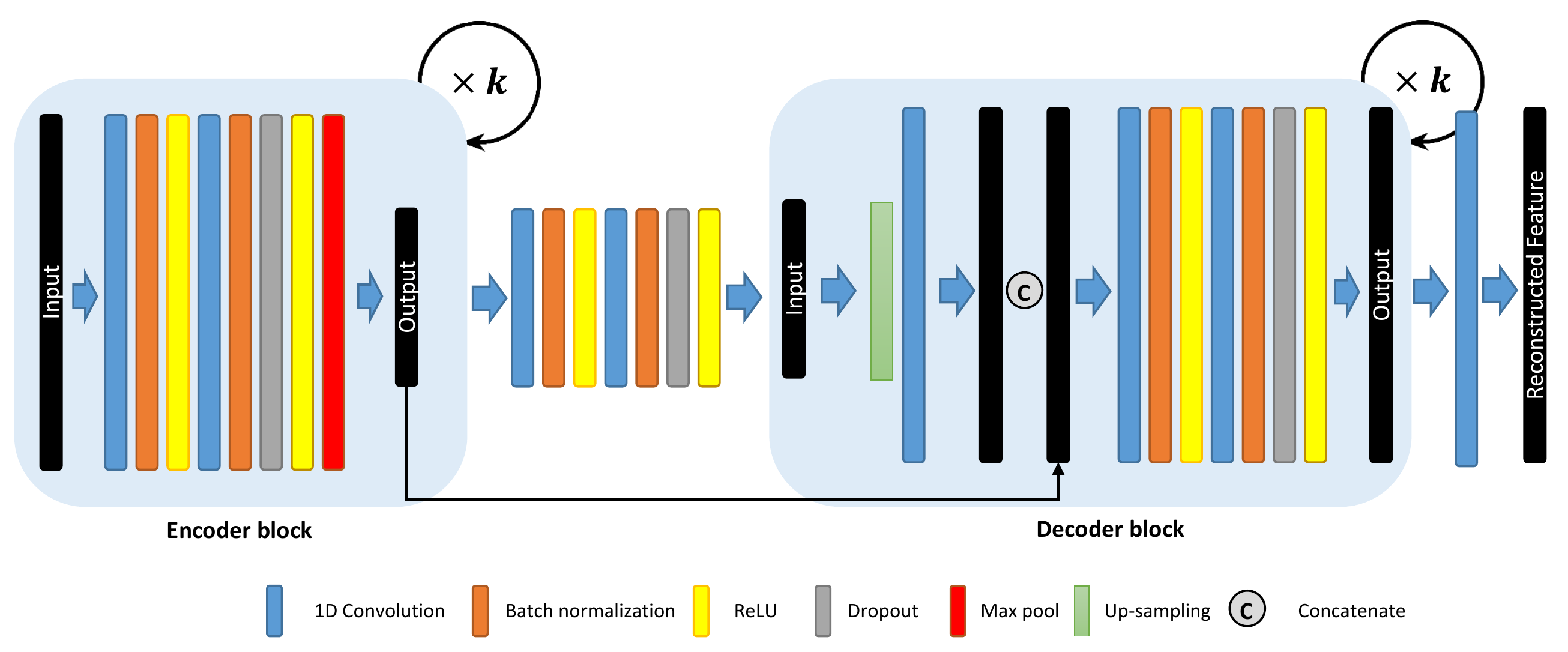}
		\caption{Structure of the generator based on U-Net.}
		\label{fig:fig_generator}
	\end{figure}

	The generator is composed of three multiscale generators as shown in Figure \ref{fig:fig_overview}. The proposed generator adopts basic concepts from U-Net \cite{ronneberger2015u} for one-dimensional(1D) time-series feature reconstruction by substituting the original 2D convolutions with 1D operations. U-Net \cite{ronneberger2015u} was invented for image segmentation, concatenating feature maps from different levels to improve segmentation performance and it combines low-level detail information with high-level semantic information. It has achieved promising performance on various image segmentation problems. Based on CNN, the U-Net architecture composes of an encoder, a decoder, and skipped connections between the encoder and the decoder. The encoder is composed of several levels of convolutional operations, which extract increasingly abstract representations of the input data, and a down sampling block to reduce input complexity. The convolutional operations followed by max-pooling down sampling applied to encode the input image into feature representations at multiple different levels. The decoder block is also composed of several levels of convolutional operations, up sampling blocks, and concatenation blocks. The decoder block semantically projects the discriminative features learned by the encoder onto higher resolution image space to get a dense classification. Unlike the general usage of U-Net for image segmentation task, we found that the structure of the U-Net enables the network to consider the deep and the shallow features at the same time and improve the effectiveness of the feature extraction. Thus, it helps to reconstruct the features to satisfy the conditions we designed. Whereas Gaussian noise is input to the generator in the general GAN, we provide noise only in the form of dropout applied on several layers of the generator. The architecture of the generators is shown in Figure \ref{fig:fig_generator}. The generator consists of the encoder and the decoder which process the input data. 
	
	Even though the U-Net enables the network to consider the deep and the shallow features at the same time, the U-Net with deep levels of convolutional operations usually focuses on more localized features and the detail information of input can be lost at the higher feature levels. Therefore, the integration of different feature levels may lead to a better performance of the feature extraction. The three multiscale generators have a different number of encoder and decoder blocks: 1) three encoder-decoder blocks (G3), 2) four encoder-decoder blocks (G4), 3) five encoder-decoder blocks (G5). The proposed multiscale generators could use two encoder-decoder blocks (G2) and five encoder-decoder blocks (G5). However, the G2 structure is too shallow to extract the correct features and the G6 is too heavy while generating the same result as G5. 
	
	\subsubsection{Discriminator} 
	
\begin{figure}[!t]
		\centering
		\includegraphics[width=\columnwidth]{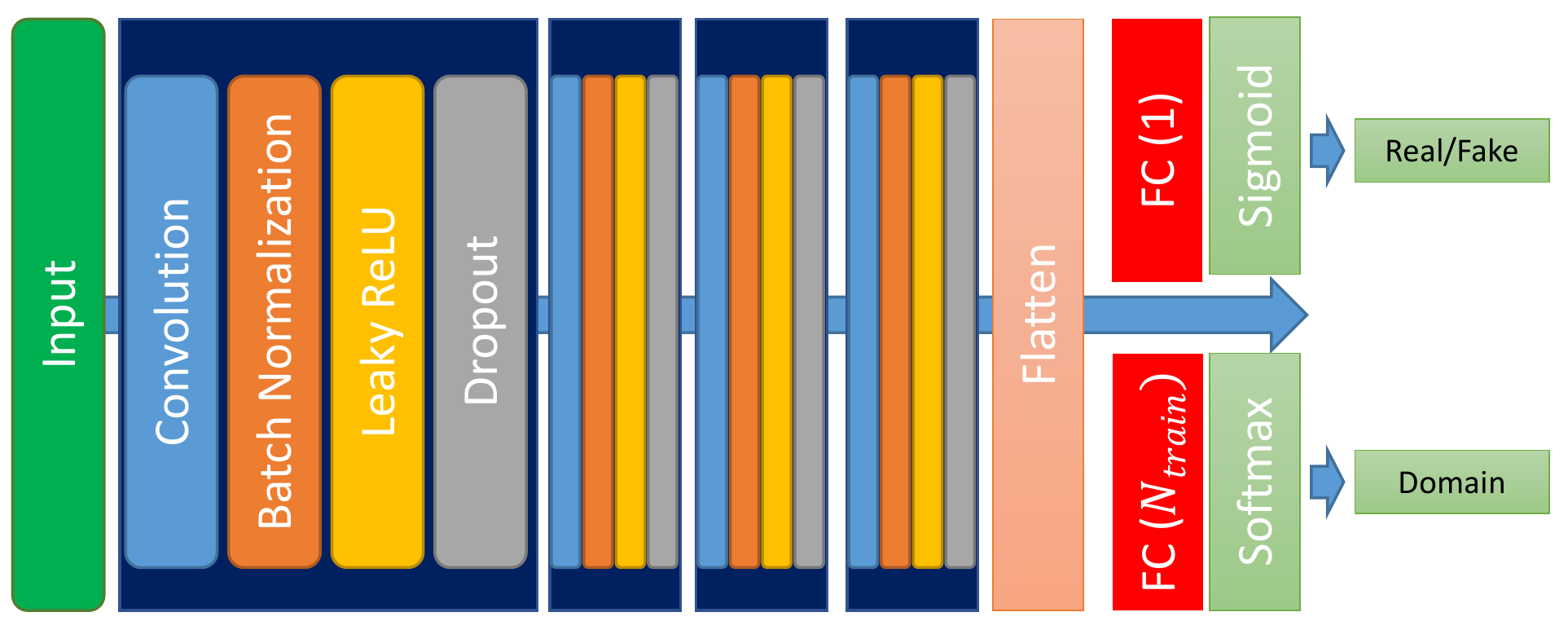}
		\caption{Structure of the discriminator.}
		\label{fig:fig_discriminator}
	\end{figure}

	The proposed GAN was developed for purpose of domain adaptation from multiple sources. Generally, the discriminator is trained to distinguish generated and real data in GAN. However, our discriminator also has a functionality of domain discriminator which distinguishes the data domain. It is because the proposed framework for the generalized multiscale feature extraction aims to extract the features of bearing wear from multiple domains. In this study, the data domain implies what bearings and under what conditions they were collected. The adversarial training enables the generator to extract high-level features containing domain-unbiased information, and makes it hard for the discriminator to classify only real or fake (denoted as a degree) as well as source domain. To do so, the convolutional layer of discriminator extract the features of input (reconstructed features of real and fake data from the generator) with a Leaky ReLU, and then the two separated linear layers output the true degree and the domain classification, respectively (shown in Figure \ref{fig:fig_discriminator}).
	
	\subsubsection{Classifier}
	
	As mentioned in Section \ref{introduction}, it is necessary to determine the HS of the machinery for accurate RUL prediction, rather than a regression model is trained directly for RUL prediction, because the machine in a healthy state would not have any remarkable difference in the run-to-failure bearing wear. To improve the performance of the HS division and the RUL prediction, we employ the structure of the classifier network to the independent network in the proposed GAN structure. In Figure \ref{fig:fig_overview}, the first stage involves a generalized multiscale feature extraction $G_{HS}$ for the HS division and the HS division model $C_{RUL}$ with the CNN and the transformed NSP images from the extracted features. In the second stage, a generalized multiscale feature extraction model $G_{RUL}$ for the RUL prediction is trained and a CNN-LSTM model $C_{RUL}$ predicts the RUL with the transformed NSP images from the extracted features.
	
	\subsubsection{Loss functions and training procedure}
	
	In this study, it is assumed that the vibration signals are collected by two high-frequency vibration sensors placed horizontally and vertically on each bearing, and a number of $N_{train}$ run-to-failure vibration data in the whole life cycle from different bearings can be used to train the proposed DNNs. Suppose that $X_{j}=\{x_j^i\}^{n_j}_{i=1}\in \mathcal{R}^{N_{samples}}, j=1,2,...,N_{train}$ denotes $n_j$ sequential training samples from the $j$th bearing, where $N_{samples}$ is the dimension of the sample. In each bearing data, two samples exist: each from horizontal and vertical vibration sensors.
	
	We aim to extract generalized prognostic features to improve the performance of the HS division and the RUL prediction. In the HS division, labeling of the target data is necessary for supervised learning to distinguish the features of the HS of a machine under different bearing degradation conditions. The training dataset can be divided into two health stages based on the time of acquisition. The initial part, corresponding to the nominal functioning in the entire vibration dataset, is labeled as healthy, and the last part of the sample duration, when the bearing is damaged, is labeled as unhealthy. This is on the assumption that degrading data of rotating machinery are obtained until the point of failure. This manner  not only reduces the effort of labeling but also improves the prediction of HS division when no ground truth of HS is available (most of the open dataset has no information about HS). The corresponding HS labels are expressed as follows.
	\begin{equation}
	\label{eq:HSlabels}
	{y_{HS}}_j^i=\left\{
	\begin{array}{@{}ll@{}}
	0, & \text{if}\ i<n_j\times p \\
	1, & \text{if}\ i>n_j\times (1-p)
	\end{array}\right.
	\end{equation} 
	where ${y_{HS}}_j^i$ represents the HS label of $x_j^i$ and $p$ is the percentage of the total degradation process. 
	The percentage of the total degradation process is used for labeling the first portion $p$ of the total degradation process as healthy and the last portion $p$ as unhealthy. The assumption is labeling a small portion of the run-to-failure dataset where the signals are clearly distinguishable into healthy and unhealthy when no labels for health stages are given. In the previous study \cite{suh2020supervised}, we found that the FPT is not sensitive to $p$ ranging from 0.05 to 0.5. $p=0.05$ is used in the experiments in Section \ref{sec:experimentalresults}. By the CNN-based binary regression model, the HS and the FPT of the $j$th bearing, ${T_{FPT}}_j$, can be determined. 

	Next, the RUL labels are expressed as follows.
	\begin{equation}
	\label{eq:RULlabels}
	{y_{RUL}}_j^i=\frac{n_j - i}{n_j - {T_{FPT}}_j}
	\end{equation}
	where ${y_{RUL}}_j^i$ denotes the RUL label of $x_j^i$. 
	
	
	The generator network is decomposed into two sub-generators: $G_{HS}$ and $G_{RUL}$. The two sub-generators correspond to discriminators $D_{HS}$ and $D_{RUL}$ with the same structure. For the stability of the training procedure and the quality of the generated data, we apply the WGAN-GP to the objective function to guide the training process. The objective functions are defined as follows.
	
	\begin{equation}
	\label{eq:dloss}
	\begin{split}
	\mathop{\mathbb{L}_D}&(x,d;\theta_D) = - \mathop{\mathbb{E}_x}[D_R(x)] + \mathop{\mathbb{E}_{x}}[D_R(G(x))] + \\
	&\alpha \mathop{\mathbb{E}_{\hat{x}\sim P_{\hat{x}}}}[(\Arrowvert \nabla_{\hat{x}} D_R(\hat{x}) \Arrowvert_2 -1 )^2] + \lambda_{D}\mathop{\mathbb{L}_{CE}}(D(x), d),
	\end{split}
	\end{equation}
	
	\begin{equation}
	\label{eq:gloss}
	\begin{split}
	\mathop{\mathbb{L}_G}(x,d,y;\theta_G) = &\mathop{\mathbb{E}_{x}}[D_R(G(x))] + \lambda_{G} \mathop{\mathbb{L}_{C}}(G(x), y) \\
	&- \lambda_{D}\mathop{\mathbb{L}_{CE}}(D_D(x), d),
	\end{split}
	\end{equation}
	\begin{equation}
	\label{eq:closs}
	\begin{split}
	\mathop{\mathbb{L}_C}(x,y;\theta_C)=\left\{
	\begin{array}{@{}ll@{}}
	\mathop{\mathbb{L}_{C_{HS}}}(G(x),y), & \text{for the HS division}\\
	\mathop{\mathbb{L}_{C_{RUL}}}(G(x),y), & \text{for the RUL prediction}\\
	\end{array}\right.
	\end{split}
	\end{equation}
	where $x$ is a sequential training sample, $y$ is the HS label in the case of the HS division and the RUL label in the case of the RUL prediction, $d$ is the associated label of the domain, $\lambda_D$ is a hyperparameter that controls the effect of domain generalization, $\lambda_G$ controls the relative importance of different loss terms, $\mathop{\mathbb{L}_{CE}}$ denotes the standard cross-entropy loss function, and $\theta_D$, $\theta_G$, and $\theta_C$ are parameters of the discriminator, the generator, and the classifier, respectively. Both the real data and generated data are entered into the discriminator, while the data reality and the domain classification are the output of the discriminator $D_R$ and $D_D$, respectively. Whereas the discriminator $D$ is trained to minimize $\mathop{\mathbb{L}_D}$ for distinguishing the data from different domains and between the real and the generated data simultaneously, the generator $G$ is trained to minimize $\mathop{\mathbb{L}_G}$. Furthermore, the classifier $C$, which is the CNN-based HS division $C_{HS}$ or the LSTM-based RUL predictor $C_{RUL}$, is trained to classify the HS or predict the RUL. In other words, the generator reconstructs the generalized features to fool the discriminator and to improve the performance of the HS division and the RUL prediction. Additionally, by reconstructing features which are similar to the real data, feature values can be generated within a certain range, and the time-series properties can be maintained. Finally, the classifier losses for the HS division and the RUL prediction are defined as follows.
	
	\begin{equation}
	\label{eq:closs2}
	\begin{split}
	\mathop{\mathbb{L}_{C_{HS}}}(G(x),y) = &\mathop{\mathbb{L}_{BCE}}(G(x),y),\\
	\mathop{\mathbb{L}_{C_{RUL}}}(G(x),y)=&\lambda_{MAE}\mathop{\mathbb{L}_{MAE}}(G(x),y) \\& + \lambda_{RMSE}\mathop{\mathbb{L}_{RMSE}}(G(x),y) 
	\\&+ \lambda_{MAPE}\mathop{\mathbb{L}_{MAPE}}(G(x),y)
	\end{split}
	\end{equation}
	where $\mathop{\mathbb{L}_{BCE}}$ denotes the standard binary cross-entropy loss, $\mathop{\mathbb{L}_{MAE}}$ is mean absolute error (MAE) in (\ref{eq:MAE}), $\mathop{\mathbb{L}_{RMSE}}$ is root mean square error (RMSE) in (\ref{eq:RMSE}), and $\mathop{\mathbb{L}_{MAPE}}$ is mean absolute percentage error (MAPE) in (\ref{eq:MAPE}).

	Three different reconstructed features are transformed into three channels of an NSP image by the following two steps. The first step is to compress the reconstructed two-channel features into nested clusters. Each channel feature is mapped onto the x- and y-axis and three different reconstructed features from the three multiscale generators are colored in red, green, and blue, respectively. To represent the intensity of each cluster from the reconstructed features, the mapped values count was translated into pixel intensity. In the second step, three scatter plots are aggregated together to form a single RGB image as shown in Figure \ref{fig:fig_overview}.
	
	The training details for the proposed multiscale feature extraction method are summarized in Algorithm \ref{Algo}. 
	
	
	\begin{algorithm}[!t]
		\caption{Training procedure for generalized multiscale feature extraction using adversarial networks. We use default values of $\alpha = 10$, $\lambda_D=1$, $\lambda_G=50$}\label{Algo}
		\begin{algorithmic}[1]
			\REQUIRE Batch size $m$, Adam hyperparameters $\eta$, hyperparameter for $\lambda_D$, $\lambda_G$.
			\STATE \textbf{Input:} $\{x_j^i\}^{n_j}_{i=1}\in \mathcal{R}^{N_{samples}\times 2}, j=1,2,...,N_{train}$, $y$, $d$.
			\FOR{number of training iterations}
			\STATE Sample $\{x_j^{i}\}^m_{i=1}$ a batch from the training dataset, corresponding HS or RUL label $\{y_j^i\}^m_{i=1}$, and corresponding domain label $\{d^i\}^m_{i=1}$.
			\FOR{$k=$3 to 5}
			\STATE Update discriminator $D$ by descending the gradient of (\ref{eq:dloss}) with $G_k$:
			\STATE $\theta_D \gets \theta_D - \eta_D \nabla_{\theta_D} \mathop{\mathbb{L}_D}(x,d;\theta_D)$
			\STATE Update generator $G_k$ by descending the gradient of (\ref{eq:gloss}):
			\STATE $\theta_{G_k} \gets \theta_{G_k} - \eta_G \nabla_{\theta_{G_k}} \mathop{\mathbb{L}_G}(x,d,y;\theta_{G_k})$
			\STATE Update classifier $C$ by descending the gradient of (\ref{eq:closs}) and (\ref{eq:closs2}) with $G_k$:
			\STATE $\theta_C \gets \theta_C - \eta_C \nabla_{\theta_C} \mathop{\mathbb{L}_C}(x,y;\theta_C)$
			\ENDFOR
			\ENDFOR
			\FOR{$k=$ 3 to 5}
			\STATE Sample $\{x_j^{i}\}^m_{i=1}$ a batch from the dataset
			\STATE Transform $G_k(x)$ into nested clusters
			\ENDFOR
			\STATE Merge three scatter plots to a single RGB image
		\end{algorithmic}
	\end{algorithm}
	
	\subsection{Health Stage Division and Remaining Useful Life Prediction with NSP}
	
\begin{figure*}[!t]
		\centering
		\subfigure[CNN-HS]{\includegraphics[width=\columnwidth]{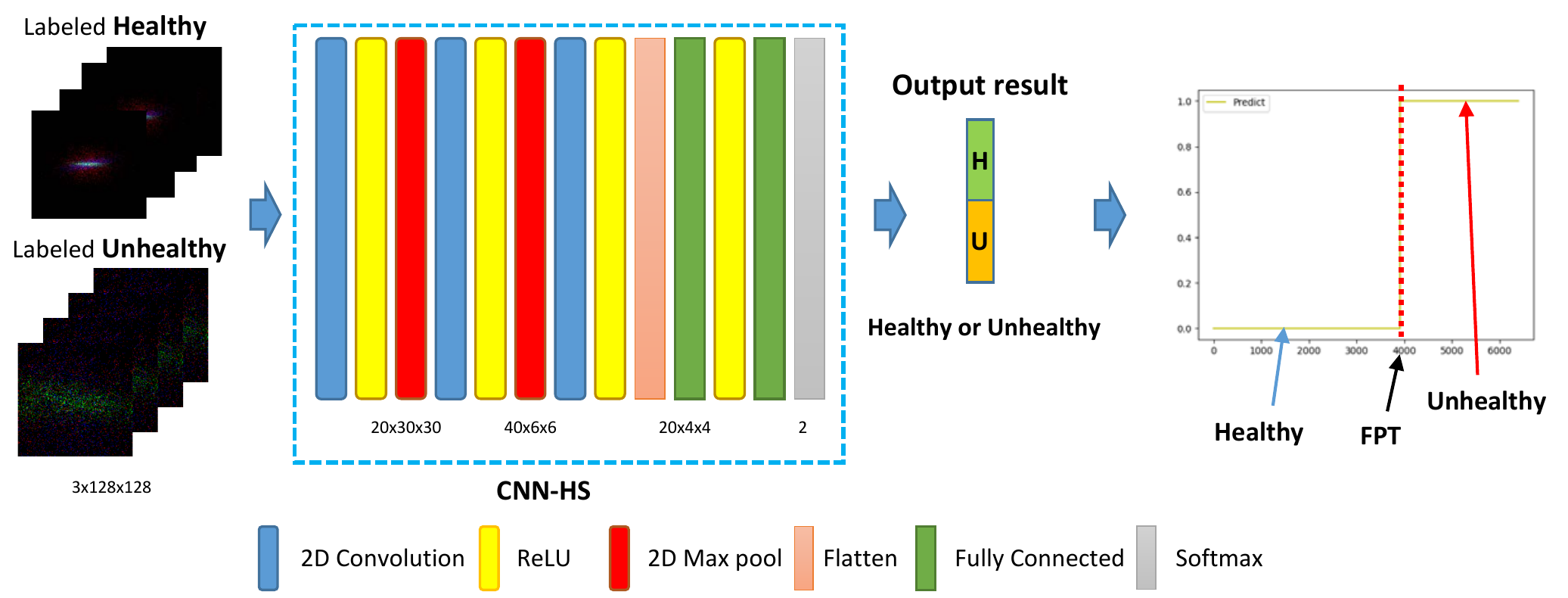}%
			\label{fig_CNNHS}}
		\hfil
		\subfigure[CNN-LSTM]{\includegraphics[width=\columnwidth]{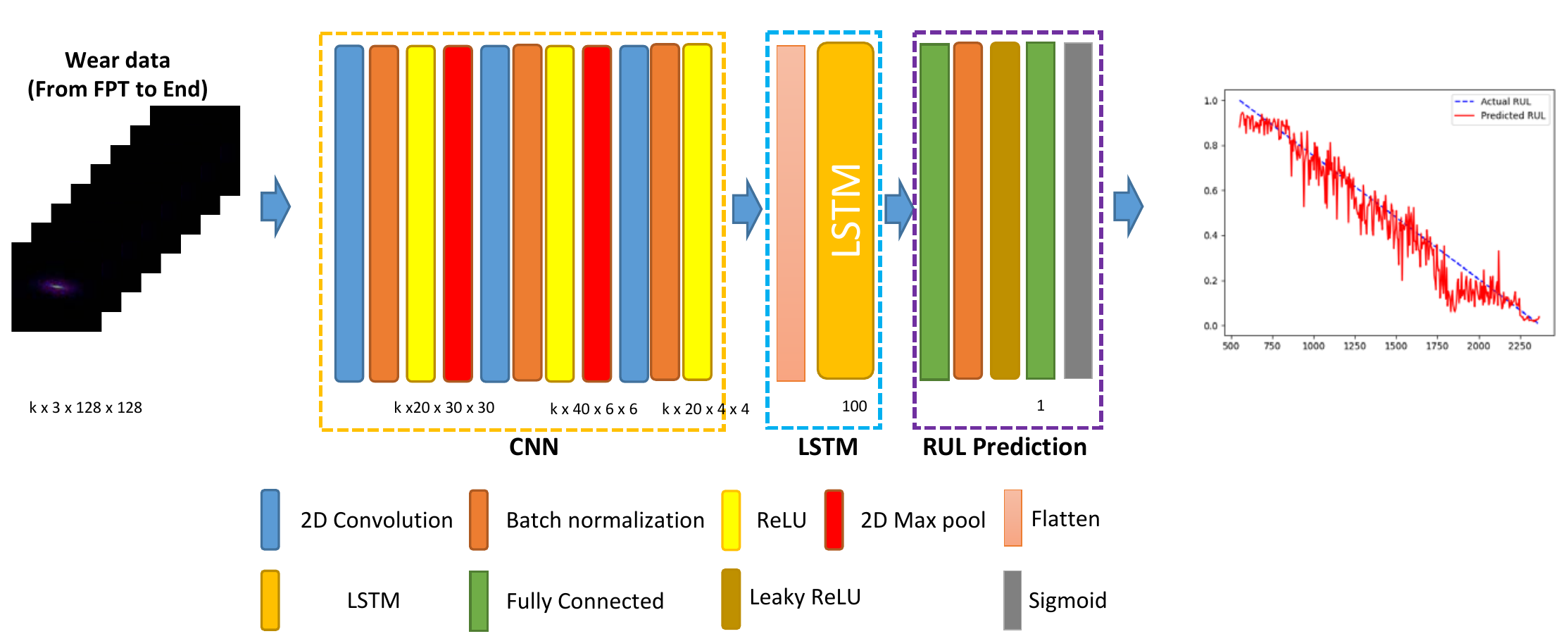}%
			\label{fig_CNNLSTM}}
		\caption{Proposed deep neural network architecture, (a) the proposed CNN-HS, (b) the proposed CNN-LSTM-RUL}
		\label{fig:fig_classifier}
		\vspace{-5mm}
	\end{figure*}

	By reconstructing continuous raw vibration time-series data to multiscale features and transforming the features into NSP images, we can change the signal processing problem to an image classification problem and classify the images by using the CNN for the HS division model (CNN-HS) \cite{suh2020supervised} and the CNN-LSTM for the RUL prediction model (CNN-LSTM-RUL). The structures of the proposed CNN-HS and CNN-LSTM-RUL are presented in Fig \ref{fig:fig_classifier}. 
	
	The trained CNN-HS and the trained CNN-LSTM can classify the NSP of an external test dataset, which was not used in training but operated under the same conditions as the rest of the training datasets. For the HS division, the binary regression results of the whole run-to-failure are computed by the CNN-HS, wherein specifying a threshold value is not necessary. As the trained CNN-HS learns the differences in the features of the NSP for healthy and unhealthy data in the training datasets, it can recognize the degradation pattern of all of the data. After determining the FPT using the CNN-HS, extracting the multiscale features, and transforming the features into the NSP images, the CNN-LSTM can recognize the degradation patterns of all of the data from the FPT to the end time. The CNN in the CNN-LSTM perform feature extraction from the transformed NSP images, and the LSTM and the RUL prediction part calculates the percentage of the RUL.

	\section{Experimental results}
	\label{sec:experimentalresults}
	\subsection{Datasets and Evaluation Metrics}
\begin{figure}[!t]
		\centering
		\subfigure[]{\includegraphics[width=\columnwidth]{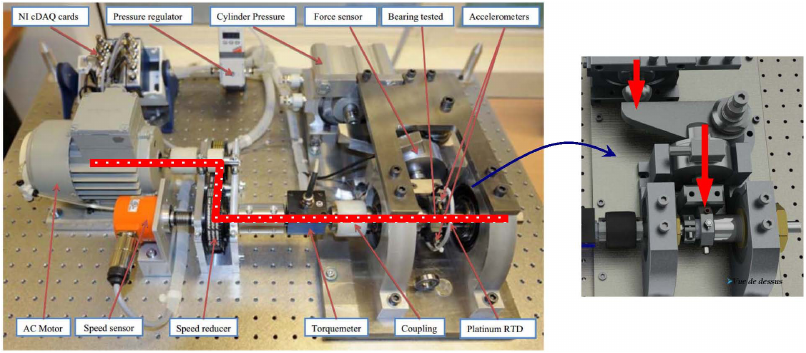}%
			\label{fig:PRONOSTIA}}
		\hfil
		\subfigure[]{\includegraphics[width=\columnwidth]{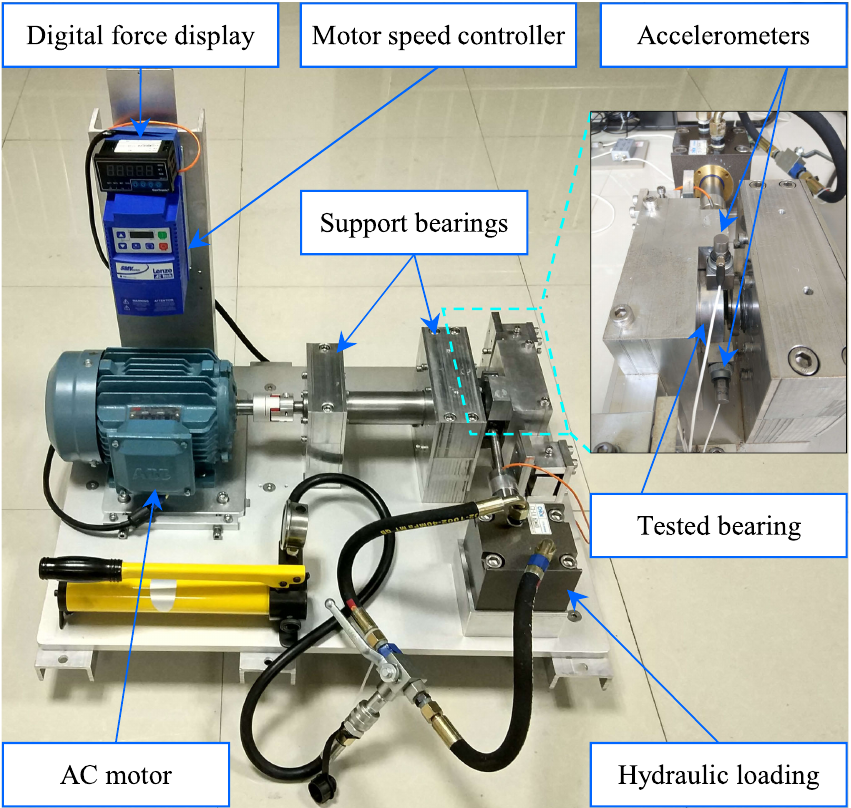}%
			\label{fig:XJTUSY}}
		\caption{Bearing test rigs (a) of the PRONOSTIA in the FEMTO dataset \cite{javed2013feature}, (b) in the XJTU-SY dataset \cite{wang2018hybrid}.}
		\label{fig:dataset}
		\vspace{-8mm}
	\end{figure}

	To evaluate the proposed method (denoted as GMFE) on the run-to-failure vibration dataset, two types of popular datasets were used: FEMTO \cite{javed2013feature} and XJTU-SY dataset \cite{wang2018hybrid}. The FEMTO dataset was collected by the PRONOSTIA test rig and has been available to the public since the IEEE PHM 2012 Prognostic Challenge (PHM 2012). The test rig mainly contains an asynchronous motor, a shaft, a speed controller, an assembly of two pulleys, and tested rolling ball bearings, which is shown in Figure \ref{fig:PRONOSTIA}. The vibration data in the horizontal direction were investigated. The dataset was composed of 17 run-to-failure data in which a single bearing was tested (two columns of vibration data: horizontal and vertical). The sampling frequency of data acquisition is 25.6 KHz and the sampling time length was 0.1 s, which was recorded per 10 s. The 17 datasets are grouped into three sections by operating conditions that differ in rotation speed and radial load. Each dataset has a different run-to-failure time, requiring fault detection methods that are adaptable to time-varying operational conditions and environments. When the amplitude of the vibration data exceeds 20 g, the run-to-failure experiments were stopped and the bearing is considered defective. More details about the test rig and the datasets can be referred from \cite{javed2013feature}. The XJTU-SY dataset contains complete run-to-failure data of 15 rolling element bearings that were collected by the Xi'an Jiaotong University and the Changxing Sumyoung Technology \cite{wang2018hybrid}. The vibration signals were collected by two accelerometers placed at 90 degrees on the housing of the tested bearings, as shown in Figure \ref{fig:XJTUSY}. The sampling frequency is 25.6 KHz and the sampling time length was 1.28 s, which was recorded per minute. Like the FEMTO dataset, the accelerated degradation bearings tests are stopped when the amplitude of the vibration data exceeds 20 g. 
	
	In the experiment, 14 run-to-failure datasets under two different operating conditions and 10 run-to-failure datasets under two different operating conditions are used from the FEMTO dataset and the XJTU-SY dataset, respectively. The dataset information for each experiment condition is presented in Table \ref{tab:datasetinfo}.
	
	\begin{table*}
		\caption{Dataset information}
		\label{tab:datasetinfo}
		\centering
		\begin{tabular}{ccccc}
			\hline
			Dataset & Condition & Load (N) & Speed (rpm) & The number of bearings \\
			\hline
			FEMTO & 1 & 4000 & 1800 & 7 \\
			FEMTO & 2 & 4200 & 1650 & 7 \\
			XJTU-SY & 1 & 12000 & 2100 & 5 \\
			XJTU-SY & 2 & 11000 & 2250 & 5 \\
			\hline
		\end{tabular}
	\end{table*}

	To evaluate and compare the performance of the proposed method with other methods, we adopted three evaluation metrics, which are used in various RUL prediction methods: MAE, RMSE, and MAPE which are calculated as follows.
	
	\begin{equation}
	\label{eq:MAE}
	MAE = \frac{1}{N_S} \sum_{i=1}^{N_S}\arrowvert ActRUL_i - PreRUL_i\arrowvert
	\end{equation}
	\begin{equation}
	\label{eq:RMSE}
	RMSE = \sqrt{\frac{1}{N_S} \sum_{i=1}^{N_S}(ActRUL_i - PreRUL_i)^2}
	\end{equation}
	\begin{equation}
	\label{eq:MAPE}
	MAPE = \frac{1}{N_S} \sum_{i=1}^{N_S} \frac{\arrowvert ActRUL_i - PreRUL_i \arrowvert}{ActRUL_i}
	\end{equation}
	where $N_S$ is the number of testing samples, and $ActRUL_i$ and $PreRUL_i$ are the actual RUL and the predicted RUL values corresponding to the $i$th testing sample, respectively.

	\subsection{Implementation Details}
	The experiments were all implemented using Python scripts in the PyTorch framework. Training procedures were conducted in the Linux system with four NVIDIA Quadro RTX 8000 GPUs. Our parameter settings for the HS division and the RUL prediction were the same except for $\lambda_G$: $\lambda_G=50$ for the HS division and $\lambda_G=1$, $\lambda_{MAE}=100$, $\lambda_{RMSE}=50$, $\lambda_{MAPE}=20$ for the RUL prediction. Through various testing, it is observed that the parameters mentioned in the paper yield the best performance. We chose the Adam optimizer with a learning rate of $\eta_C=\eta_G=\eta_D=1 \times 10^{-4}$, $\beta_1=0.5$, and $\beta_2=0.99$.  The batch size was 40 and the sequence lengths in the CNN-LSTM model for the FEMTO and the XJTU-SY datasets are 5 and 4, respectively.
	
	\subsection{Results}
	
	\begin{figure}[!t]
	    \centering
		\subfigure[]{\includegraphics[width=\columnwidth]{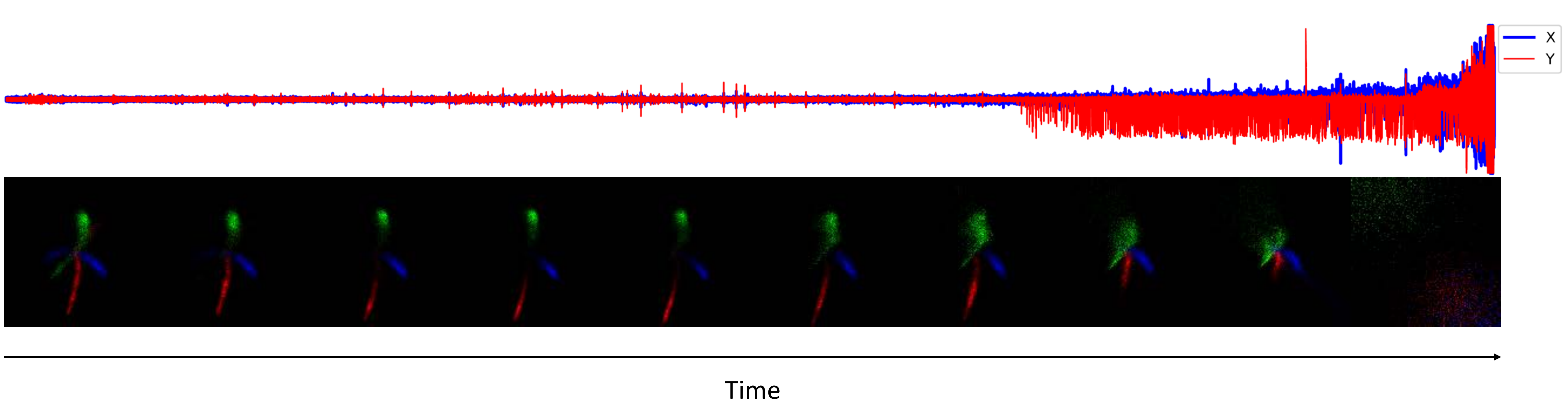}}
		\hfil
		\subfigure[]{\includegraphics[width=\columnwidth]{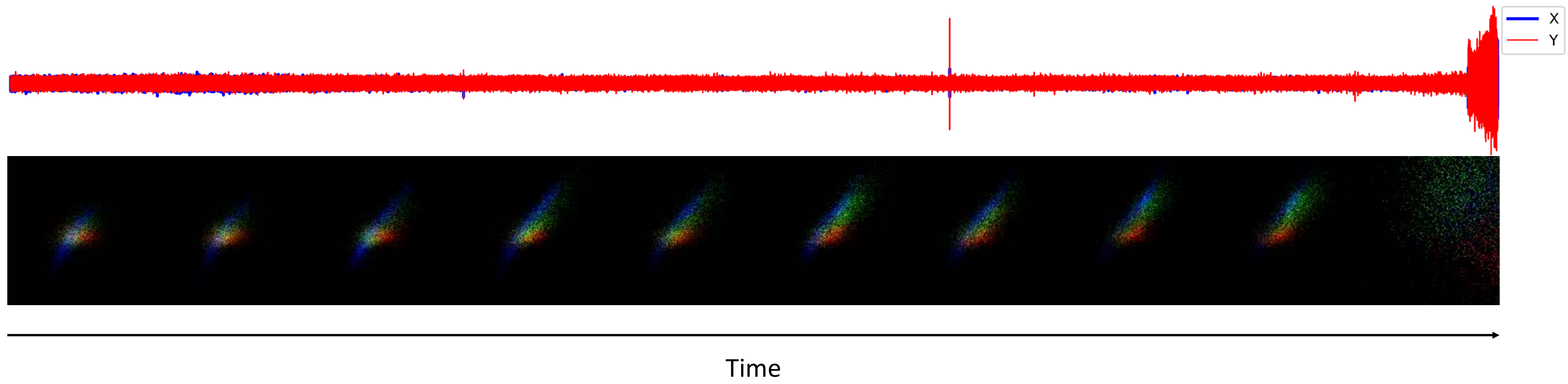}}
		\hfil
		\subfigure[]{\includegraphics[width=\columnwidth]{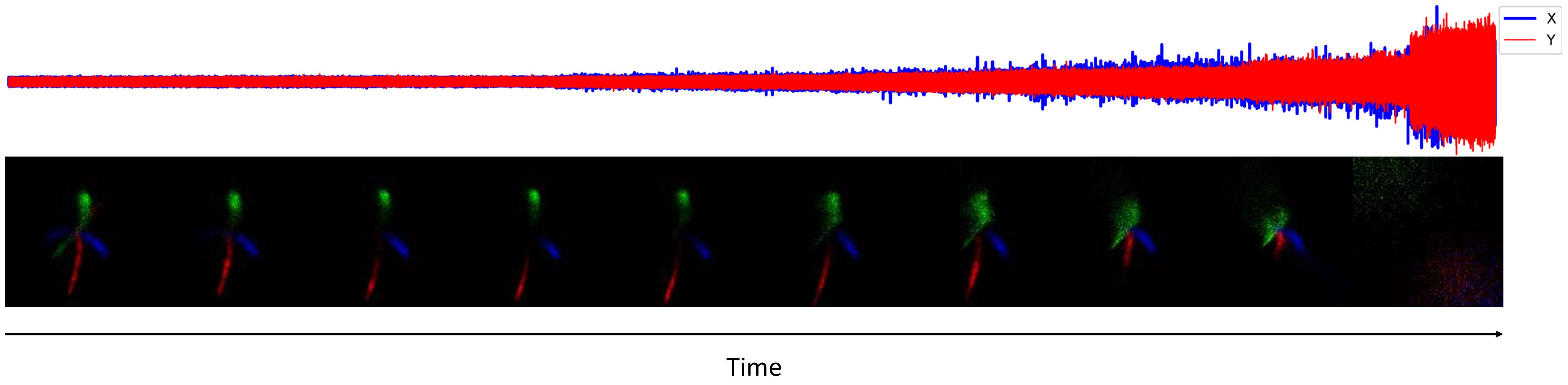}}
		\hfil
		\subfigure[]{\includegraphics[width=\columnwidth]{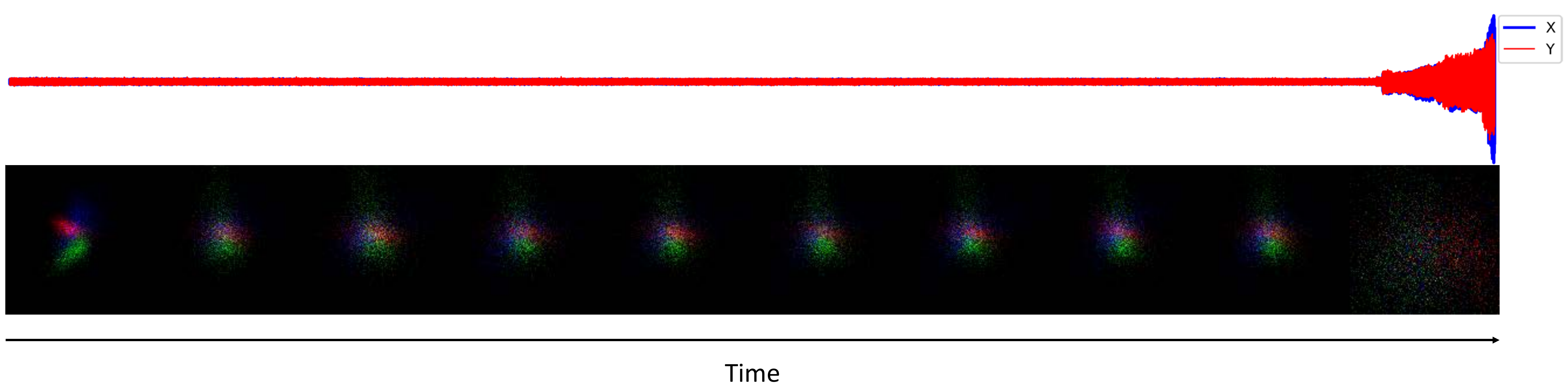}}
		\caption{Examples of transformed NSP images on four testing bearings. (a) Bearing 1-3 in the FEMTO dataset, (b) Bearing 2-6 in the FEMTO dataset, (c) Bearing 1-3 in the XJTU-SY dataset, (d) Bearing 2-1 in the XJTU-SY dataset.}
		\label{fig:NSP_examples}
	\end{figure}
	
	Figure \ref{fig:NSP_examples} shows the examples of transformed NSP images on four example testing bearings from the FEMTO and the XJTU-SY datasets. The transformed NSP images can efficiently represent the health of machinery by extracting useful features from raw vibration data. 
	
	\begin{table*}[!t]
		\caption{Evaluation of RUL prediction on the FEMTO and XJTU-SY datasets. }
		\label{tab:comparison}
		\centering
		\resizebox{\linewidth}{!}{
			\begin{tabular}{cccccc}
				\hline
				Methods & Metric & FEMTO Condition1 & FEMTO Condition2 & XJTU-SY Condition 1 & XJTU-SY Condition 2\\
				\hline
				\multirow{3}{*}{MCNN \cite{li2019deep}} 
				& MAE 	& 0.1856 $\pm$ 0.0546 & 0.1738 $\pm$ 0.0521 & 0.2274 $\pm$ 0.0496 & 0.2063 $\pm$ 0.0427\\
				& RMSE 	& 0.2344 $\pm$ 0.0701 & 0.2105 $\pm$ 0.0557 & 0.2484 $\pm$ 0.0452 & 0.2302 $\pm$ 0.0319\\
				& MAPE 	& 0.8943 $\pm$ 0.4802 & 1.3567 $\pm$ 0.7913 & 1.1659 $\pm$ 0.3749 & 1.1802 $\pm$ 0.3945\\
				\hline
				\multirow{3}{*}{DANN \cite{li2020data}} 
				& MAE 	& 0.1739 $\pm$ 0.0542 & 0.1771 $\pm$ 0.0515 & 0.2588 $\pm$ 0.0499 & 0.2112 $\pm$ 0.0891\\
				& RMSE 	& 0.2063 $\pm$ 0.0603 & 0.2060 $\pm$ 0.0578 & 0.2966 $\pm$ 0.0587 & 0.2398 $\pm$ 0.0973\\
				& MAPE 	& \textbf{0.4633 $\pm$ 0.0811} & \textbf{0.4970 $\pm$ 0.1200} & \textbf{0.4340 $\pm$ 0.0816} & \textbf{0.3872 $\pm$ 0.1535}\\
				\hline
				\multirow{3}{*}{CNN-HS \cite{suh2020supervised}} 
				& MAE 	& 0.1697 $\pm$ 0.0267 & 0.2694 $\pm$ 0.0913 & 0.1919 $\pm$ 0.0469 & 0.1238 $\pm$ 0.0434\\
				& RMSE 	& 0.1959 $\pm$ 0.0297 & 0.2932 $\pm$ 0.0890 & 0.2172 $\pm$ 0.0505 & 0.1429 $\pm$ 0.0419 \\
				& MAPE 	& 0.7983 $\pm$ 0.4179 & 1.2953 $\pm$ 0.9574 & 0.6574 $\pm$ 0.3538 & 0.6563 $\pm$ 0.4435\\
				\hline
				\multirow{3}{*}{GMFE (ours)}
				& MAE 	& \textbf{0.0906 $\pm$ 0.0237} & \textbf{0.1644 $\pm$ 0.0456} & \textbf{0.1620 $\pm$ 0.0476} & \textbf{0.1132 $\pm$ 0.0476}\\
				& RMSE 	& \textbf{0.1053 $\pm$ 0.0222} & \textbf{0.1870 $\pm$ 0.0408} & \textbf{0.1795 $\pm$ 0.0488} & \textbf{0.1259 $\pm$ 0.0488}\\
				& MAPE 	& 0.4794 $\pm$ 0.0859 & 0.8515 $\pm$ 0.5413 & 0.5306 $\pm$ 0.1445 & 0.5721 $\pm$ 0.4197\\	
				\hline
				\hline
				\multirow{3}{*}{No FPT} 
				& MAE 	& 0.0678 $\pm$ 0.0212 & 0.2142 $\pm$ 0.0591 & 0.2541 $\pm$ 0.0802 & 0.2145 $\pm$ 0.0790\\
				& RMSE 	& 0.0842 $\pm$ 0.0274 & 0.2317 $\pm$ 0.0540 & 0.2647 $\pm$ 0.0770 & 0.2217 $\pm$ 0.0747\\
				& MAPE 	& 0.5860 $\pm$ 0.3584 & 1.3122 $\pm$ 1.0572 & 2.7105 $\pm$ 3.3264 & 0.9537 $\pm$ 0.6268\\		
				\hline
				\multirow{3}{*}{1D-GAN}
				& MAE 	& 0.1983 $\pm$ 0.0944 & 0.2879 $\pm$ 0.1059 & 0.2393 $\pm$ 0.1239 & 0.1295 $\pm$ 0.0664\\
				& RMSE 	& 0.2099 $\pm$ 0.0936 & 0.3143 $\pm$ 0.1098 & 0.2529 $\pm$ 0.1200 & 0.1427 $\pm$ 0.0650\\
				& MAPE 	& 0.8234 $\pm$ 0.4407 & 1.5198 $\pm$ 1.3135 & 0.6782 $\pm$ 0.2530 & 1.2928 $\pm$ 1.6532\\		
				\hline
		\end{tabular}}
	\end{table*}
	
	To evaluate the proposed method, while one bearing dataset is used for testing, the other bearings under the same operating conditions are adopted to train the DNNs. The experimental results are averaged by five trials to reduce the effect of model randomness. To address the advantage of the proposed method, we compare the proposed method with the multiscale CNN method (MCNN) by Li et al. \cite{li2019deep}, the deep adversarial neural networks (DANN) by Li et al. \cite{li2020data}, and the proposed CNN-LSTM method with the supervised HS division method \cite{suh2020supervised}. We implemented MCNN to compare to the proposed method by following the detailed experimental setting in \cite{li2019deep} because the XJTU-SY dataset was not used in \cite{li2019deep}.
	Table \ref{tab:comparison} shows the quantitative evaluation results comparing to other methods by using three evaluation metrics. 
	It is observed that the proposed method showed much lower prediction errors than others in terms of MAE and RMSE, while the MAPE of the proposed method is slightly higher than DANN \cite{li2020data}. When testing the various parameter setting, we found that more weight on MAE and RMSE than MAPE in the loss function yields the best performance of RUL prediction in all the ranges. If the weight on MAPE increases, the tendency of the predicted RUL decreases, and the overall performance is degraded. It is the reason why $\lambda_{MAPE}=20$ is a lower value than $\lambda_{MAE}=100$ and $\lambda_{RMSE}=50$ in the best parameter setting for loss function.

	To evaluate the effect of FPT and NSP, the ablation study is conducted. ‘No FPT’ is adopted and the degradation is considered to start at the beginning of machine operation to indicate the benefits of the HS division and FPT determination. The ‘1D-GAN’ method is implemented without merging the multiscale features by transforming into the NSP image and the proposed CNN-LSTM-RUL to show the advantage of the transformed NSP images and the CNN-LSTM-RUL. The proposed method compared to ‘No FPT’ and ‘1D-GAN’ showed better results, indicating that the proposed HS division method is effective in capturing machine degradation at the beginning point and the CNN-LSTM-RUL with the transformed NSP images enhances the generalization ability on various datasets.
	
	\begin{figure}[!t]
		\centering
		\subfigure[]{\includegraphics[width=0.45\columnwidth]{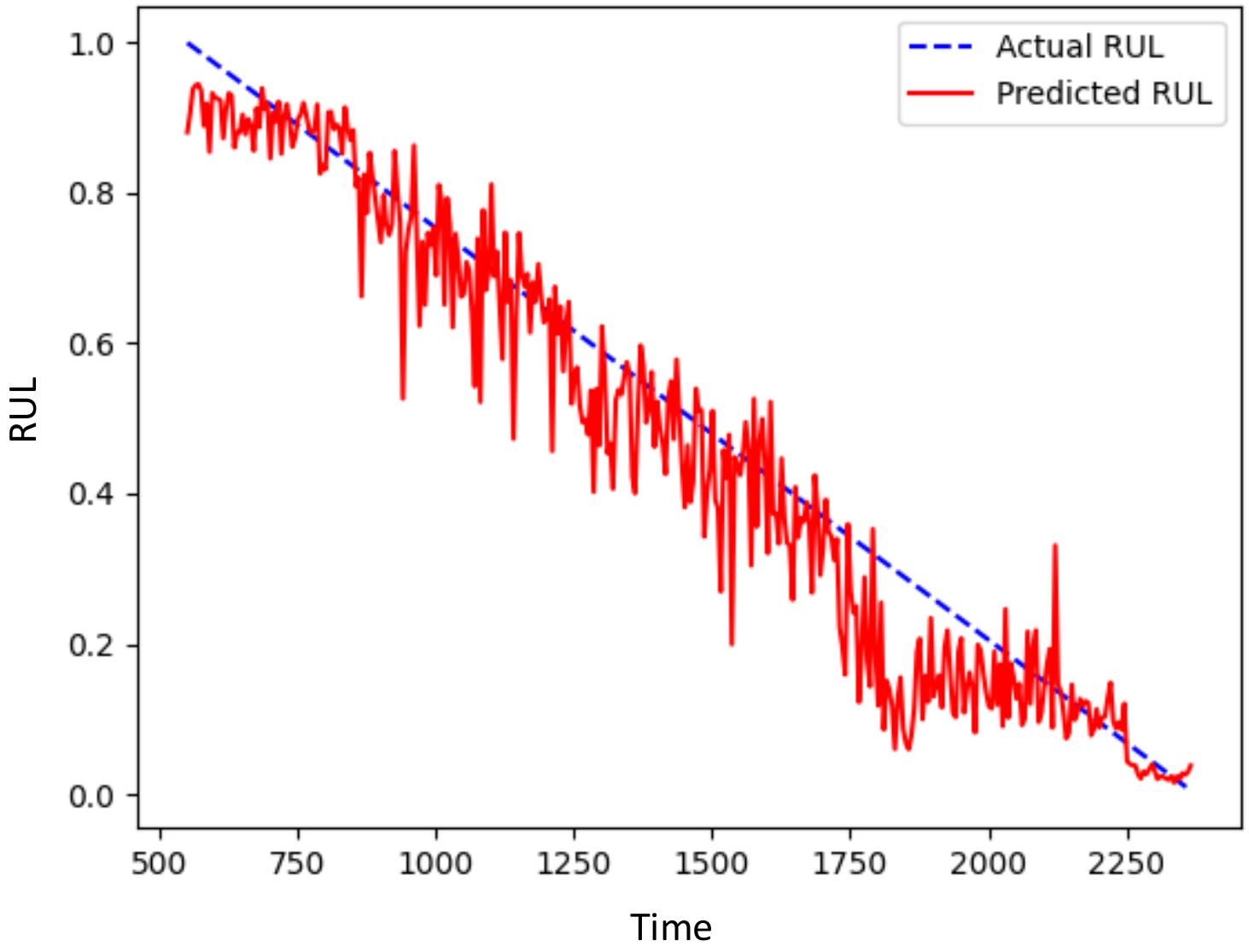}%
			\label{fig_FEMTO1_3}}
		\hfil
		\subfigure[]{\includegraphics[width=0.45\columnwidth]{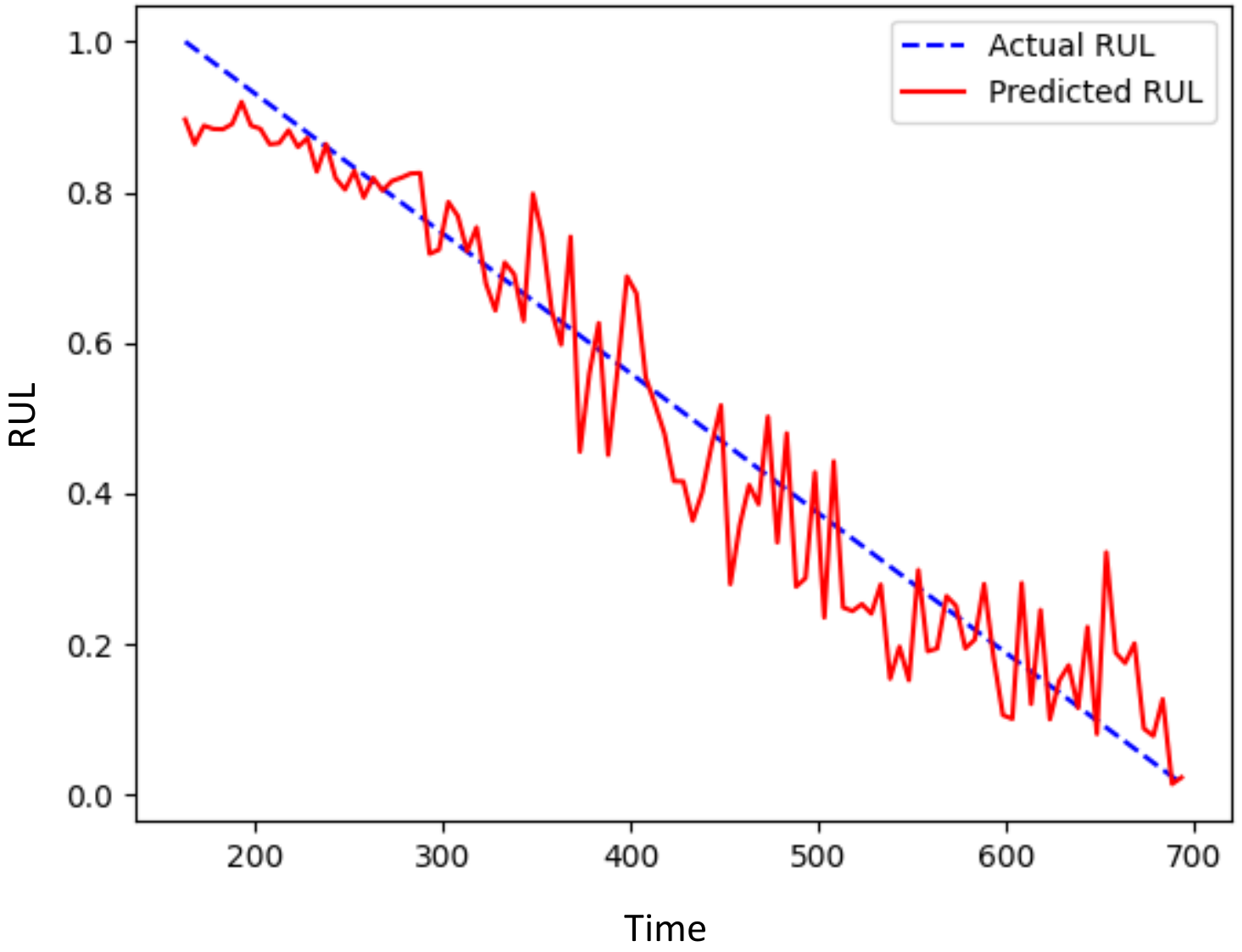}%
			\label{fig_FEMTO2_6}}
		\hfil
		\subfigure[]{\includegraphics[width=0.45\columnwidth]{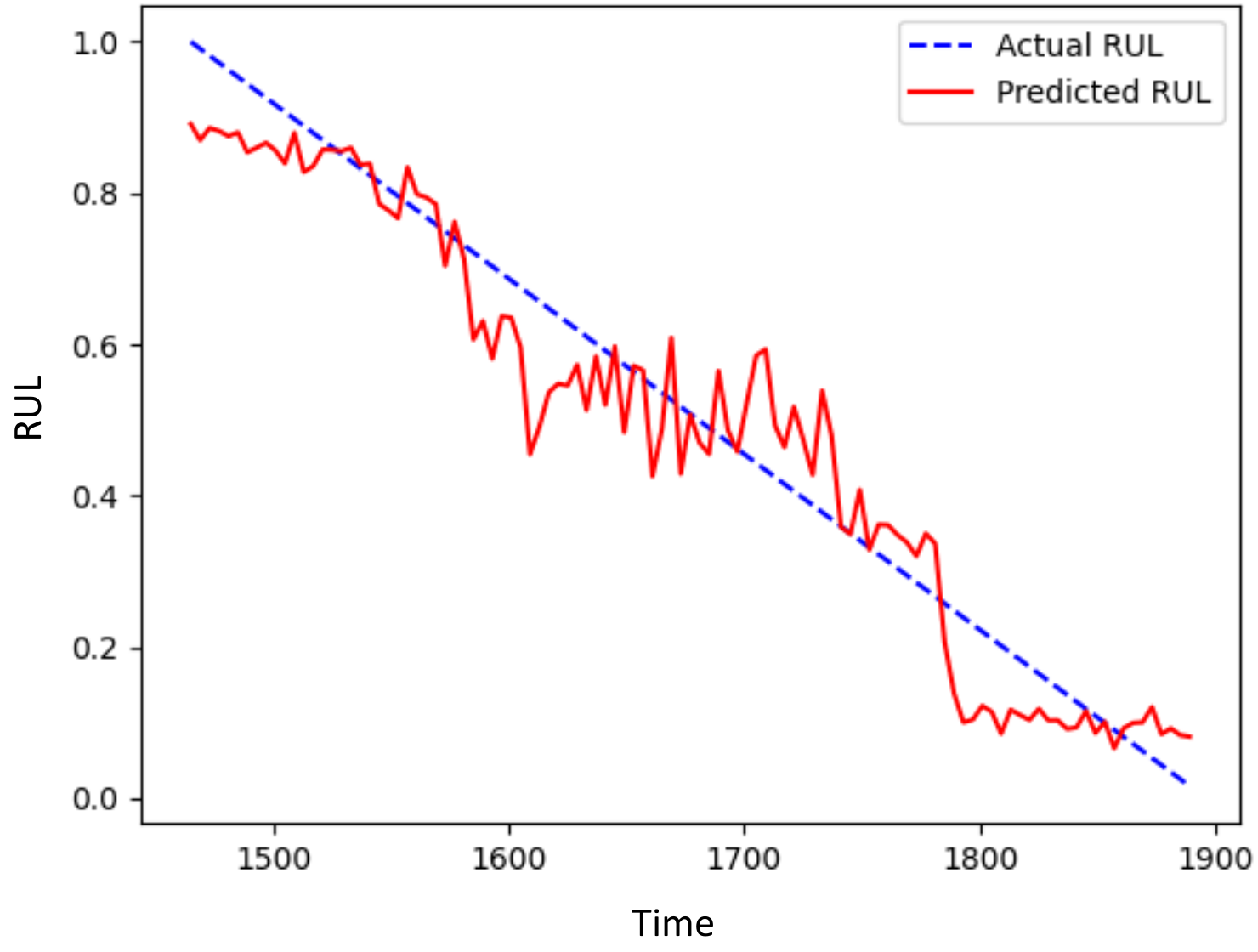}%
			\label{fig_XJTU1_3}}
		\hfil
		\subfigure[]{\includegraphics[width=0.45\columnwidth]{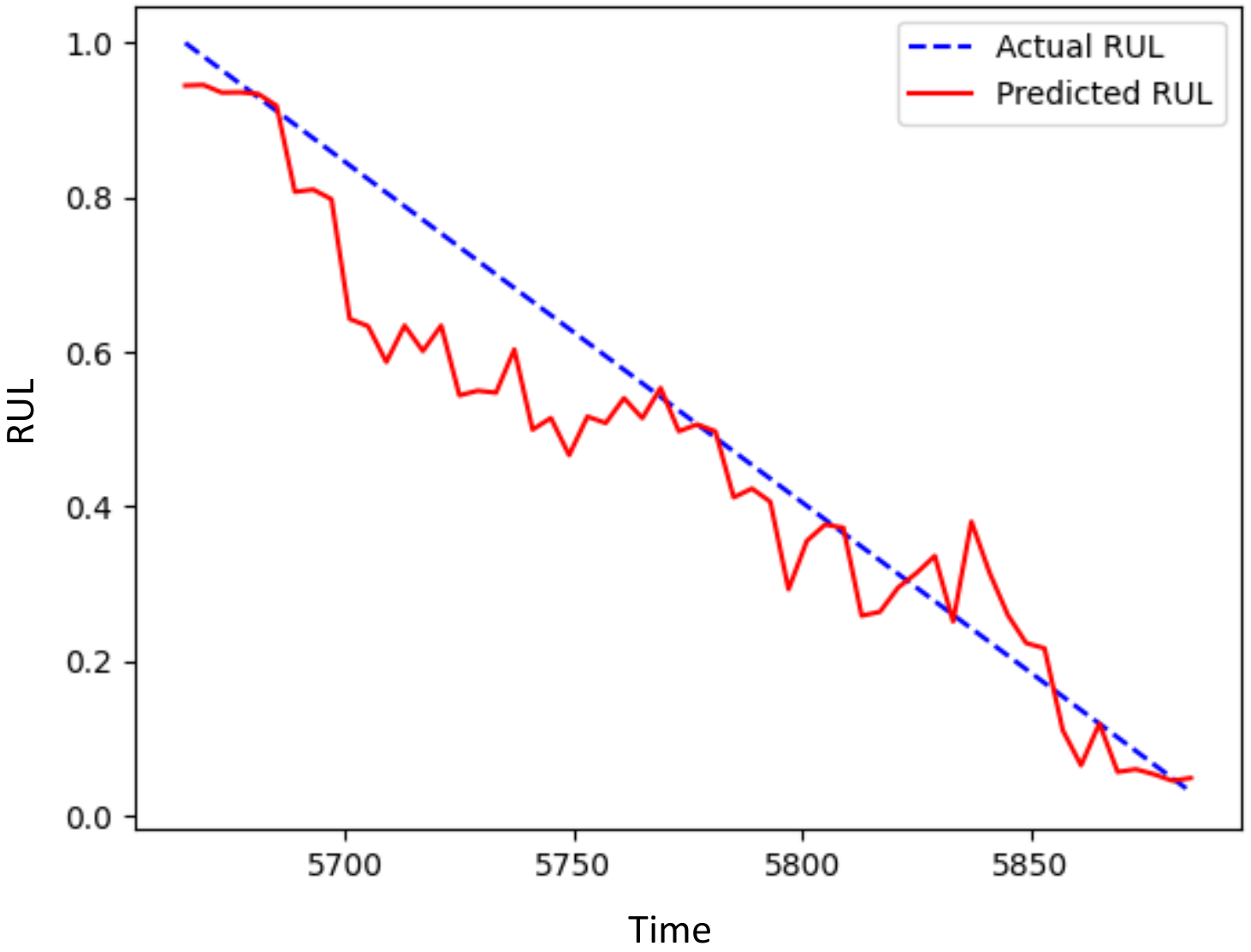}%
			\label{fig_XJTU2_1}}
		\caption{RUL prediction results on four testing bearings. (a) Bearing 1-3 in the FEMTO dataset, (b) Bearing 2-6 in the FEMTO dataset, (c) Bearing 1-3 in the XJTU-SY dataset, (d) Bearing 2-1 in the XJTU-SY dataset. }
		\label{fig:RUL_prediction}
		\vspace{-5mm}
	\end{figure}
	
	Figure \ref{fig:RUL_prediction} presents the RUL prediction results on four testing bearings in the two datasets. The degradation evaluations start when the FPTs of the testing bearings are detected. It can be observed that the degrading patterns are effectively captured by the proposed method and the errors between the actual RUL and the predicted RUL are quite small.

	\section{Conclusion}
	\label{sec:conclusion}
	In this study, we have proposed generalized multiscale feature extraction and RUL prediction method using multiscale GAN. To extract generalized prognostic features, we designed a multiscale 1D U-Net architecture for the generator and adversarial learning procedure between the generator and the discriminator which contains the reality and domain discriminator. The adversarial learning procedure learns the distributions of multiple training data from different bearings and extracts domain-invariant generalized prognostic features. The proposed method does not require any domain knowledge and manual setting to extract generalized features and predict the RUL. A CNN-based binary regression model for the determination of the HS without any threshold and a CNN-LSTM model for the RUL prediction have been proposed, which can predict the RUL with fewer errors and higher prognosis accuracy than other existing methods.
	The experimental results indicated that the proposed method, where CNN and CNN-LSTM are combined with NSP based on the reconstructed features of bearing wear, could capture the degrading patterns effectively and achieved better RUL prediction results compared to other methods. 
	
	To further improve the performance of the proposed method, more amounts training data are required. However, annotated data in the whole life cycle of bearings are hard to collect in the real industry fields and to define the healthy condition of bearings. Further related research works have to be conducted on unsupervised bearing degradation data. In the future, we plan to extend the proposed method to fault diagnosis and prognosis of other components in manufacturing environments and further develop the fault prediction model with attention mechanism and unsupervised transfer learning.

	\section*{Acknowledgments}
	This work was supported by the KIST Europe Institutional Program (Project No. 12120).

	\bibliography{mybibfile}

\end{document}